\documentstyle[aclap,psfig,rotate]{article}

\title{Tagging Grammatical Functions}

\author{Thorsten Brants, Wojciech Skut, Brigitte Krenn\\
        Universit{\"a}t des Saarlandes\\
	Computational Linguistics\\
        D-66041 Saarbr{\"u}cken, Germany\\
        {\tt \{brants,skut,krenn\}@coli.uni-sb.de}\\[1ex]
	{\em In Proceedings of EMNLP-2, Providence, RI, 1997}}

\def\argmax{\mathop{\rm argmax}}
\def\baselinestretch{0.955}

\begin{document}

\maketitle
\bibliographystyle{fullname}

\begin{abstract}

This paper addresses issues in automated treebank construction.  We show
how standard part-of-speech tagging techniques extend to the more
general problem of structural annotation, especially for determining
grammatical functions and syntactic categories. Annotation is viewed as
an interactive process where manual and automatic processing alternate.
Efficiency and accuracy results are presented. We also discuss further
automation steps. 

\end{abstract}

\section{Introduction}

The aim of the work reported here is to construct a corpus of German
annotated with syntactic structures (treebank). The required size of
the treebank and granularity of encoded information make it necessary
to ensure high annotation efficiency and accuracy. Annotation
automation has thus become one of the central issues of the project.

In this section, we discuss the relation
between automatic and manual annotation. Section \ref{sec:scheme}
focuses on the annotation format employed in our treebank. The 
annotation software is presented in section \ref{sec:tool}. Sections
\ref{sec:tagger} and \ref{sec:tagphrase} deal with automatic
assignment of grammatical functions and phrasal categories.
Experiments on automating the annotation are presented in section
\ref{sec:experiment}.

\subsection{Automatic vs. Manual Annotation}

A problem for corpus annotation is the trade-off between efficiency,
accuracy and coverage.  Although accuracy increases significantly as
annotators gain expertise, incorrect hand-parses still occur. Their
frequency depends on the granularity of the encoded information.

Due to this residual error rate, automatic annotation of frequently occurring 
phenomena is likely to yield better results than even well-trained
human annotators. For infrequently occurring constructions, however,
manual annotation is more reliable, as is manual annotation of
phenomena involving non-syntactic information (e.g., resolution of
attachment ambiguities based on world knowledge).

As a consequence, efficiency and reliability of annotation can be
significantly increased by combining automatic annotation with human
processing skills and supervision, especially if this combination is
implemented as an interactive process.

\section{Annotation Scheme}
\label{sec:scheme}

\begin{figure*}
\hrule
\bigskip
\centerline{\psfig{file=kneipe.eps,width=14cm,rwidth=13.5cm}}
\def\h#1{\hspace*{#1}}
\centerline{\footnotesize\sf The\h{1em}dream\h{2em}of\h{2em}the\h{1.8em}small\h{2.3em}inn\h{3em}has\h{2.6em}he\h{2em}yet\h{3em}not\h{3em}given\h{.3em}up}
\smallskip
\centerline{\em `He has not yet given up the dream of a small inn.'}
\vspace{.5ex}
\hrule
\caption{Example sentence}
\label{fig:ex1}
\end{figure*}

Existing treebanks of English (\cite{Marcusea:94}, \cite{Sampson95}, 
\cite{Black96}) contain conventional phrase-structure trees augmented
 with annotations for discontinuous constituents. As this encoding
 strategy is not well-suited to a free word order language like
 German, we have focussed on a less surface-oriented level of
 description, most closely related to the LFG f-structure, and
 representations used in dependency grammar.  To avoid confusion with
 theory-specific constructs, we use the generic term {\em argument
 structure} to refer to our annotation format. The main advantages of
 the model are: it is relatively theory-independent and closely
 related to semantics. For more details on the linguistic
 specifications of the annotation scheme see \cite{Skutea97}.  A
similar approach has been also successfully applied in the TSNLP
database, cf. \cite{LehmannEA:96}.

In contrast to conventional phrase-structure grammars, argument
structure annotations are not influenced by word order. Local and
non-local dependencies are represented in the same way, the latter 
indicated by crossing branches in the hierarchical structure, as
shown in figure \ref{fig:ex1} where in the VP the terminals of the direct
object {\sf OA} ({\it den Traum von der kleinen Gastst\"atte}) are not
adjacent to the head {\sf HD} {\it
aufgegeben}\footnote{See appendix \ref{sec:tagsets} for a description
of tags used throughout this paper.}. For a related handling of
non-projective phenomena see \cite{TapanainenJaervinen:97}.

Such a representation permits clear separation of word order (in the
surface string) and syntactic dependencies (in the hierarchical
structure). Thus we avoid explicit explanatory statements about the
complex interrelation between word order and syntactic structure in
free word order languages. Such statements are generally
theory-specific and therefore are not appropriate for a descriptive
approach to annotation. The relation between syntactic dependencies
and surface order can nontheless be inferred from the data. This
provides a promising way of handling free word order phenomena.%
\footnote{`Free' word order is a function of several interacting 
parameters such as category, case and topic-focus
articulation. Varying the order of words in a sentence yields a
continuum of grammaticality judgments rather than a simple right-wrong
distinction.  }.

\section{Annotation Tool}
\label{sec:tool}

Since syntactic annotation of corpora is time-consuming, a partially
automated annotation tool has been developed in order to increase
efficiency.

\subsection{The User Interface}

For optimal human-machine interaction, the tool supports immediate
graphical representation of the structure being annotated.

Since keyboard input is most efficient for assigning categories to
words and phrases, cf. \cite{LehmannEA:96,Marcusea:94}, and structural
manipulations are executed most efficiently using the mouse, both an
elaborate keyboard and optical interface is provided.  As suggested by
Robert MacIntyre\footnote{personal communication, Oct. 1996}, it is
most efficient to use one hand for structural commands with the mouse
and the other hand for short keyboard input.

By additionally offering online menus for commands and labels, the
tool suits beginners as well as experienced users. Commands such as
``group words'', ``group phrases'', ``ungroup'', ``change labels'',
``re-attach nodes'', ``generate postscript output'', etc. are
available.

The three tagsets (word, phrase, and edge labels) used by the annotation
tool are variable. They are stored together with the corpus, which
allows easy modification and exchange of tagsets.  In addition,
appropriateness checks are performed automatically. Comments can be
added to structures.

Figure \ref{FigScreen} shows a screen dump of the graphical interface. 

\begin{figure*}[t]
\begin{center}
\hrule
\bigskip
\centerline{\psfig{file=screen.eps,width=\textwidth}}
\bigskip
\hrule  
\mbox{} 
\caption{Screen dump of the annotation tool}
\label{FigScreen}
\end{center}
\end{figure*}

\subsection{Automating Annotation}

Existing treebank annotation tools are characterised by a high degree
of automation. The task of the annotator is to correct the output of a
parser, i.e., to eliminate wrong readings, complete partial parses,
and adjust partially incorrect ones.

Since broad-coverage parsers for German, especially robust parsers
that assign predicate-argument structure and allow crossing branches,
are not available, or require an annotated traing corpus 
(cf. \cite{Collins96}, \cite{Eisner96}).

As a consequence, we have adopted a bootstrapping approach, and
gradually increased the degree of automation using already annotated
sentences as training material for a stochastic processing module.

This aspect of the work has led to a new model of human supervision.
Here automatic annotation and human supervision
are combined interactively whereby annotators are asked to confirm
the {\em local} predictions of the parser. The size of such `supervision
increments' varies from local trees of depth one to larger chunks,
depending on the amount of training data available.

We distinguish six degrees of automation:

\begin{enumerate}
\item[0)] Completely manual annotation.
\item[1)] The user determines phrase boundaries and syntactic categories (S, NP,
	VP, \dots). The program automatically assigns
	grammatical functions. The annotator can alter the assigned tags
	(cf. figure \ref{fig:kanten}).
\item[2)] The user only determines the components of a new phrase (local tree
	of depth 1), while both category and function labels are
	assigned automatically. Again, the annotator has the option of
	altering the assigned tags (cf. figure \ref{fig:knoten}).
\item[3)] The user selects a substring and a category, whereas the
	entire structure covering the substring is determined
	automatically (cf.\ figure \ref{fig:subphrase}).
\item[4)] The program performs simple bracketing, i.e., finds `kernel
	phrases' without the user having to explicitly mark phrase
	boundaries. The task can be performed by a chunk parser that
	is equipped with an appropriate finite state grammar
	\cite{Abney96}.
\item[5)] The program suggests partial or complete parses.
\end{enumerate}

A set of 500 manually annotated training sentences (step 0) was
sufficient for a statistical tagger to reliably assign grammatical
functions, provided the user determines the elements of a phrase and
its category (step 1). Approximately 700 additional sentences have
been annotated this way. Annotation efficiency increased by 25 \%,
namely from an average annotation time of 4 minutes to 3 minutes per
sentence (300 to 400 words per hour). The 1,200
sentences were used to train the tagger for automation step
2. Together with improvements in the user interface, this increased
the efficiency by another 33\%, from approximately 3 to 2 minutes (600
words per hour). The fastest annotators cover up to 1000 words per
hour.

At present, the treebank comprises 3000 sentences, each annotated
independently by two annotators. 1,200 of the sentences are compared
with the corresponding second annotation and are cleaned, 1,800 are
currently cleaned.

In the following sections, the automation steps 1 and 2 are presented in
detail.

\begin{figure}
\hrule
\bigskip
\centerline{\psfig{file=kanten.ps,width=\columnwidth,angle=-90}}
\centerline{\footnotesize\em `the bonus program for frequent fliers starting in 1993'}
\medskip
\hrule  
\caption{Example for automation level 1: the user has marked {\em das},
the {\sf AP}, {\em Bonusprogramm}, and the {\sf PP} as a constituent
of category {\sf NP}, and the tool's task is to determine the new edge
labels (marked with question marks), which are, from left to right,
{\sf NK, NK, NK, MNR}.}
\label{fig:kanten}
\end{figure}

\begin{figure}
\hrule
\bigskip
\centerline{\psfig{file=knoten.ps,width=\columnwidth,angle=-90}}
\centerline{\footnotesize\em `the bonus program for frequent fliers starting in 1993'}
\medskip
\hrule  
\caption{Example for automation level 2: the user has marked {\em das},
the {\sf AP}, {\em Bonusprogramm} and the {\sf PP} as a
constituent, and the tool's task is to determine the new node and
edge labels (marked with question marks).}
\label{fig:knoten}
\end{figure}

\begin{figure}
\hrule
\bigskip
\centerline{\psfig{file=subphrase.ps,width=\columnwidth,angle=-90}}
\centerline{\footnotesize\em `the bonus program for frequent fliers starting in 1993'}
\medskip
\hrule  
\caption{Example for automation level 3: the user has marked the words
as a constituent, and the tool's task is to determine simple
sub-phrases (the {\sf AP} and {\sf PP}) as well as the new node and
edge labels (cf. previous figures for the resulting structure).}
\label{fig:subphrase}
\end{figure}

\section{Tagging Grammatical Functions}
\label{sec:tagger}

\subsection{The Tagger}

In contrast to a standard part-of-speech tagger which estimates lexical and 
contextual probabilities of tags from sequences of
word-tag pairs in a corpus, (e.g. \cite{Cutting92,Feldweg95}), the tagger for grammatical functions works with
lexical and contextual probability measures $P_Q(\cdot)$ depending on
the category of the mother node ($Q$). Each phrasal category ({\sf S,
VP, NP, PP} etc.) is represented by a different Markov model. The
categories of the daughter nodes correspond to the outputs of the Markov
model, while grammatical functions correspond to states.

The structure of a sample sentence is shown in figure \ref{fig:ex}.
Figure \ref{fig:markov} shows those parts of the Markov models for
sentences ({\sf S}) and verb phrases ({\sf VP}) that represent the
correct paths for the example.\footnote{cf. appendix \ref{sec:tagsets}
for a description of tags used in the example}

\begin{figure}
\hrule
\bigskip
\psfig{file=peterbes.eps,angle=-90,width=8.5cm,rwidth=8cm}
\vspace*{-3ex}
\def\h#1{\hspace*{#1}}
{\sf\h{.3em}himself\h{1.4em}visited\h{2em}has\h{2.1em}Peter\h{1.3em}Sabine\h{1.1em}never}\\[1ex]
\centerline{\em `Peter never visited Sabine himself'}
\vspace{.3ex}
\hrule
\caption{Example sentence}
\label{fig:ex}
\end{figure}

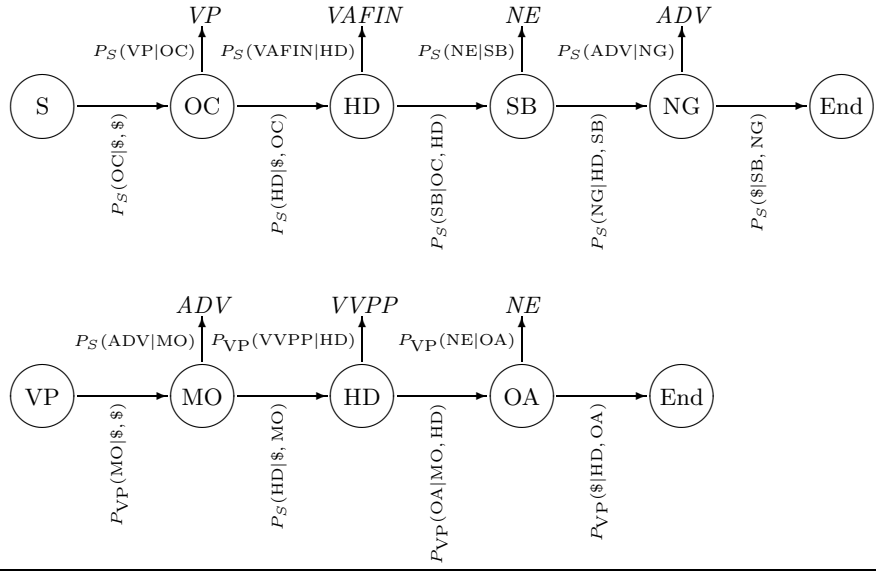
\begin{figure*}[t]
\hrule

\begin{center}
\def\kreis(#1,#2)#3{\put(#1,#2){\circle{10}\makebox(0,0){#3}}}
\small
\setlength{\unitlength}{.85mm}
\let\probsize=\tiny
\let\statesize=\small
\def\g#1{\mbox{\probsize #1}}
\def\gt#1{\mbox{\probsize #1}}
\def\gs#1{\mbox{\statesize #1}}
\begin{picture}(145,43)(15,-25)
	\kreis(25,0){\gs{S}}
		\put(30.5,0){\vector(1,0){14}}
		\put(37,-1){\makebox(0,0)[t]{\rotate[l]{\probsize $P_S(\g{OC}|\g{\$},\g{\$})$}}}
	\kreis(50,0){\gs{OC}}
	\put(50,13){\makebox(0,0)[b]{\small\em VP}}
	\put(50,5.5){\vector(0,1){7}}
	\put(49,8.5){\makebox(0,0)[r]{\probsize$P_S(\gt{VP}|\g{OC})$}}
		\put(55.5,0){\vector(1,0){14}}
		\put(62,-1){\makebox(0,0)[t]{\rotate[l]{\probsize$P_S(\g{HD}|\g{\$},\g{OC})$}}}
	\kreis(75,0){\gs{HD}}
	\put(75,13){\makebox(0,0)[b]{\small\em VAFIN}}
	\put(75,5.5){\vector(0,1){7}}
	\put(74,8.5){\makebox(0,0)[r]{\probsize$P_S(\gt{VAFIN}|\g{HD})$}}
		\put(80.5,0){\vector(1,0){14}}
		\put(87,-1){\makebox(0,0)[t]{\rotate[l]{\probsize$P_S(\g{SB}|\g{OC},\g{HD})$}}}
	\kreis(100,0){\gs{SB}}
	\put(100,13){\makebox(0,0)[b]{\small\em NE}}
	\put(100,5.5){\vector(0,1){7}}
	\put(99,8.5){\makebox(0,0)[r]{\probsize$P_S(\gt{NE}|\g{SB})$}}
		\put(105.5,0){\vector(1,0){14}}
		\put(112,-1){\makebox(0,0)[t]{\rotate[l]{\probsize$P_S(\g{NG}|\g{HD},\g{SB})$}}}
	\kreis(125,0){\gs{NG}}
	\put(125,13){\makebox(0,0)[b]{\small\em ADV}}
	\put(125,5.5){\vector(0,1){7}}
	\put(124,8.5){\makebox(0,0)[r]{\probsize$P_S(\gt{ADV}|\g{NG})$}}
		\put(130.5,0){\vector(1,0){14}}
		\put(137,-1){\makebox(0,0)[t]{\rotate[l]{\probsize$P_S(\g{\$}|\g{SB},\g{NG})$}}}
	\kreis(150,0){\footnotesize End}
\end{picture}

\begin{picture}(95,45)(-10,-25)
	\kreis(-25,0){\gs{VP}}
		\put(-19.5,0){\vector(1,0){14}}
		\put(-13,-1){\makebox(0,0)[t]{\rotate[l]{\probsize$P_{\g{VP}}(\g{MO}|\g{\$},\g{\$})$}}}
	\kreis(0,0){\gs{MO}}
	\put(0,13){\makebox(0,0)[b]{\small\em ADV}}
	\put(0,5.5){\vector(0,1){7}}
	\put(-1,8.5){\makebox(0,0)[r]{\probsize$P_S(\gt{ADV}|\g{MO})$}}
		\put(5.5,0){\vector(1,0){14}}
		\put(12,-1){\makebox(0,0)[t]{\rotate[l]{\probsize$P_S(\g{HD}|\g{\$},\g{MO})$}}}
	\kreis(25,0){\gs{HD}}
	\put(25,13){\makebox(0,0)[b]{\small\em VVPP}}
	\put(25,5.5){\vector(0,1){7}}
	\put(24,8.5){\makebox(0,0)[r]{\probsize$P_{\g{VP}}(\gt{VVPP}|\g{HD})$}}
		\put(30.5,0){\vector(1,0){14}}
		\put(37,-1){\makebox(0,0)[t]{\rotate[l]{\probsize$P_{\g{VP}}(\g{OA}|\g{MO},\g{HD})$}}}
	\kreis(50,0){\gs{OA}}
	\put(50,13){\makebox(0,0)[b]{\small\em NE}}
	\put(50,5.5){\vector(0,1){7}}
	\put(49,8.5){\makebox(0,0)[r]{\probsize$P_{\g{VP}}(\gt{NE}|\g{OA})$}}
		\put(55.5,0){\vector(1,0){14}}
		\put(62,-1){\makebox(0,0)[t]{\rotate[l]{\probsize$P_{\g{VP}}(\g{\$}|\g{HD},\g{OA})$}}}
	\kreis(75,0){\footnotesize End}
\end{picture}
\end{center}

\hrule
\caption{Parts of the Markov models used in {\em Selbst besucht hat Peter
Sabine nie} (cf. figure \ref{fig:ex}). All unused states, transitions and outputs
are omitted.}
\label{fig:markov}
\end{figure*}  

Given a sequence of word and phrase categories $T = T_1\dots T_k$ and
a parent category $Q$, we calculate the sequence of grammatical
functions $G=G_1\dots G_k$ that link $T$ and $Q$ as

\vbox{%
\begin{eqnarray}
	\argmax_G P_Q(G|T) & & \hspace*{1em}
\end{eqnarray}
\begin{eqnarray*}
	&&= \argmax_G \frac{P_Q(G)\cdot P_Q(T|G)}{P_Q(T)}\\[1mm]
	&&= \argmax_G P_Q(G)\cdot P_Q(T|G)
\end{eqnarray*}
}

Assuming the Markov property we have

\begin{equation}
	P_Q(T|G) = \prod_{i=1}^k P_Q(T_i|G_i)
\end{equation}
and
\begin{equation}
	P_Q(G) = \prod_{i=1}^k P_Q(G_i|C_i)
\end{equation}

The contexts $C_i$ are modeled by a fixed number of surrounding
elements. Currently, we use two grammatical functions, which results in
a trigram model:

\begin{equation}
	P_Q(G) = \prod_{i=1}^k P_Q(G_i|G_{i-2},G_{i-1})
\end{equation}

The contexts are smoothed by linear interpolation of unigrams, bigrams,
and trigrams. Their weights are calculated by deleted interpolation
\cite{Brown92}.

The predictions of the tagger are correct in approx.\ 94\% of all cases.
In section \ref{sec:reliability}, we demonstrate how to cope with
wrong predictions.

\subsection{Serial Order}

As the annotation format permits trees with crossing branches, we need
a convention for determining the relative position of overlapping
sibling phrases in order to assign them a position in a Markov model.
For instance, in figure \ref{fig:ex} the range of the terminal node
positions of {\sf VP} overlaps with those of the subject {\sf SB} and
the finite verb {\sf HD}. Thus there is no single a-priori position
for the {\sf VP} node%
\footnote{Without crossing edges, the serial order of phrases is
trivial: phrase $Q_1$ precedes phrase $Q_2$ if and only if all terminal
nodes derived from $Q_1$ precede those of $Q_2$. This suffices to
uniquely determine the order of sibling nodes.}.

The position of a phrase depends on the position of its
descendants. We define the relative order of two phrases recursively
as the order of their {\em anchors}, i.e., some specified daughter
nodes. If the anchors are words, we simply take their linear order.

The exact definition of the anchor is based on linguistic knowledge. We
choose the most intuitive alternative and define the anchor as the head
of the phrase (or some equivalent function). Noun phrases do not
necessarily have a unique head; instead, we use the last element in the
noun kernel (elements of the noun kernel are determiners, adjectives,
and nouns) to mark the anchor position. Except for NPs, we employ a
default rule that takes the leftmost element as the anchor in case the
phrase has no (unique) head.

Thus the position of the {\sf VP} in figure \ref{fig:ex} is defined as
equal to the string position of {\em besucht}. The position of the
{\sf VP} node in figure \ref{fig:ex1} is equal to that of {\em
aufgegeben}, and the position of the {\sf NP} in figure
\ref{fig:kanten} is equivalent to that of {\em Bonusprogramm}.

\subsection{Reliability}
\label{sec:reliability}

Experience gained from the development of the Penn Treebank
\cite{Marcusea:94} has shown that automatic annotation is useful only if it
is absolutely correct, while wrong analyses are often difficult to
detect and their correction can be time-consuming.

To prevent the human annotator from missing errors, the tagger for
grammatical functions is equipped with a measure for the reliability
of its output.

Given a sequence of categories, the tagger calculates the most probable
sequence of grammatical functions. In addition, it computes the
probabilities of the second-best functions of each daughter node. If
some of these probabilities are close to that of the best sequence, the
alternatives are regarded as equally suited and the most probable one is
not taken to be the sole winner, the prediction is marked as unreliable
in the output of the tagger.

These unreliable predictions can be further classified in that we
distinguish ``unreliable'' sequences as opposed to ``almost
reliable'' ones.

The distance between two probabilities for the best and second-best
alternative, $p_{best}$ and $p_{second}$, is measured by their
quotient. The classification of reliability is based on thresholds.
In the current implementation we employ three degrees of reliability
which are separated by two thresholds $\theta_1$ and $\theta_2$.
$\theta_1$ separating unreliable decisions from those considered 
almost reliable. $\theta_2$ marks the difference between almost and
fully reliable predictions. 

\noindent{\bf Unreliable:}
\[
	\frac{p_{best}}{p_{second}} < \theta_1
\]
The probabilities of alternative assignments are within some small
specified distance. In this case, it is the annotator who has to
specify the grammatical function.

\noindent{\bf Almost reliable:}
\[
	\theta_1 \leq \frac{p_{best}}{p_{second}} < \theta_2
\]
The probability of an alternative is within some larger distance. In
this case, the most probable function is displayed, but the annotator
has to confirm it. 

\noindent{\bf Reliable:}
\[
\frac{p_{best}}{p_{second}} \geq \theta_2
\]
The probabilities of all alternatives are much smaller than that of
the best assignment, thus the latter is assigned.

For efficiency, an extended Viterbi algorithm is used. Instead
of keeping track of the best path only (cf.
\cite{Rabiner89}), we keep track of all paths that fall into the range
marked by the probability of the best path and $\theta_2$, i.e., we keep
track of all alternative paths with probability $p_{alt}$ for which
\[
	p_{alt} \geq \frac{p_{best}}{\theta_2}.
\]

Suitable values for $\theta_1$ and $\theta_2$ were determined
empirically (cf. section \ref{sec:experiment}).

\section{Tagging Phrase Categories}
\label{sec:tagphrase}

The second level of automation (cf. section \ref{sec:tool}) automates
the recognition of phrasal categories, and so frees the annotator from
typing phrase labels. The task is performed by an extension of the
tagger presented in the previous section where different Markov models
for each category were introduced. The annotator determines the category
of the current phrase, and the tool runs the appropriate model to
determine the edge labels.

To assign the phrase label automatically, we run all models
in parallel. Each model assigns grammatical functions and, more
important for this step, a probability to the phrase. The model
assigning the highest probability is assumed to be most adequate, and
the corresponding label is assigned to the phrase.

Formally, we calculate the phrase category $Q$ (and at the same time
the sequence of grammatical functions $G = G_1\dots G_k$) on the basis
of the sequence of daughters $T = T_1\dots T_k$ with
\[
	\argmax_Q\ \max_G P_Q(G|T).
\]

This procedure is equivalent to a different view on the same problem
involving one large (combined) Markov model that enables a very
efficient calculation of the maximum.

Let ${\cal G}_Q$ be the set of all grammatical functions that can occur
within a phrase of type $Q$. Assume that these sets are pairwise
disjoint. One can easily achieve this property by indexing all used
grammatical functions with their associated phrases and, if necessary,
duplicating labels, e.g., instead of using {\sf HD}, {\sf MO}, \dots,
use the indexed labels {\sf HD$_S$}, {\sf HD$_{VP}$}, {\sf MO$_{NP}$},
\dots This property makes it possible to determine a phrase category by
inspecting the grammatical functions involved.

When applied, the combined model assigns grammatical functions to the
elements of a phrase (not knowing its category in advance). If
transitions between states representing labels with different indices
are forced to zero probability (together with smoothing applied to other
transitions), all labels assigned to a phrase get the same index. This
uniquely identifies a phrase category.

The two additional conditions
\[
	G\in{\cal G}_{Q1} \Rightarrow G\not\in{\cal G}_{Q2}\ \ \ (Q_1
	\not= Q_2)
\]
and
\[
	G_1\in{\cal G}_Q \wedge G_2\not\in{\cal G}_Q \Rightarrow
	P(G_2|G_1) = 0
\]
are sufficient to calculate
\[
	\argmax_G P(G|T)
\]
using the Viterbi algorithm and to identify both the phrase category and
the respective grammatical functions.

Again, as described in section \ref{sec:tagger}, we calculate
probabilities for alternative candidates in order to get reliability
estimates.

The overall accuracy of this approach is approx.\ 95\%, and higher if we only
consider the reliable cases. Details about the accuracy are reported in
the next section.

\section{Experiments}
\label{sec:experiment}

To investigate the possibility of automating annotation, experiments
were performed with the cleaned part of the treebank\footnote{The
corpus is part of the German newspaper text provided on the ECI
CD-ROM. It has been part-of-speech tagged and manually corrected
previously, cf.
\cite{Schiller;Thielen:94}.} (approx. 1,200 sentences,  24,000 words).
The first run of experiments was carried out to test tagging of
grammatical functions, the second run to test tagging of phrase
categories.

\subsection{Grammatical Functions}

This experiment tested the reliability of assigning grammatical
functions given the category of the phrase and the daughter nodes
(supplied by the annotator).

Let us consider the sentence in figure \ref{fig:ex}: two sequences of
grammatical functions are to be determined, namely the grammatical
functions of the daughter nodes of S and VP. The information given for
{\em selbst besucht Sabine} is its category ({\sf VP}) and
the daughter categories: adverb ({\sf ADV}), past participle
({\sf VVPP}), and proper noun ({\sf NE}). The task is to assign the
functions modifier ({\sf MO}) to {\sf ADV}, head ({\sf HD}) to {\sf VVPP}
and direct (accusative) object ({\sf OA}) to {\sf NE}.  Similarly,
function tags are assigned to the components of the sentence ({\sf S}).

The tagger described in section \ref{sec:tagger} was used. 

The corpus was divided into two disjoint parts, one for training (90\%
of the respective corpus), and one for testing (10\%). This procedure
was repeated 10 times with different partitions. Then the average
accuracy was calculated.

\begin{table}
\caption{Levels of reliability and the percentage cases where the
tagger assigned a correct grammatical function (or would have assigned if a
decision is forced).}
\label{tab:res}
\hrule
\begin{center}
\begin{tabular}{l|rr}
	& cases		& correct \\
\hline
	& 		& 	\\[-1ex]
reliable	& 89\%	& 96.7\% \\
marked		&  7\%	& 84.3\% \\
unreliable	&  4\%	& 57.3\% \\
\hline
overall & 100\%		& 94.2\% \\
\end{tabular}
\end{center}
\hrule
\end{table}

The thresholds for search beams were set to $\theta_1=5$ and
$\theta_2=100$, i.e., a decision is classified as reliable if there is
no alternative with a probability larger than $\frac{1}{100}$ of the
best function tag. The prediction is classified as unreliable if the
probability of an alternative is larger than $\frac{1}{5}$ of the most
probable tag. If there is an alternative between these two
thresholds, the prediction is classified as almost reliable and marked
in the output (cf. section \ref{sec:reliability}: marked assignments
are to be confirmed by the annotator, unreliable assignments are
deleted, annotation is left to the annotator).

Table \ref{tab:res} shows tagging accuracy depending on the three
different levels of reliability. The results confirm the choice of
reliability measures: the lower the reliability, the lower the
accuracy.

Table \ref{tab:phrases} shows tagging accuracy depending on the
category of the phrase and the level of reliability. The table
contains the following information: the number of all mother-daughter
relations (i.e., number of words and phrases which are immediately
dominated by a mother node of a particular category), the overall
accuracy for that phrasal category and the accuraciees for the three
reliability intervals.

\begin{table}
\caption{Tagging accuracy for assigning grammatical functions depending
on the category of the mother node. For each category, the first row
shows the percentage of branches that occur within this category and the
overall accuracy, the following rows show the relative percentage and
accuracy for different levels of reliability.}
\label{tab:phrases}
\hrule
\begin{center}
\begin{tabular}{l|rr}
	& cases		& correct \\
\hline
\hline
		&	& \\[-1ex]
{\sf S} 	& 26\%	& 89.1\% \\
\hline
		&	& \\[-1ex]
decision	& 85\%	& 92.7\% \\
marked		&  8\%	& 81.9\% \\
no decision	&  7\%	& 52.9\% \\
\hline
\hline
		&	& \\[-1ex]
{\sf VP} 	& 7\%	& 90.9\% \\
\hline
		&	& \\[-1ex]
decision	& 97\%	& 92.2\%  \\
marked		&  1\%	& 57.7\% \\
no decision	&  2\%	& 52.3\% \\
\hline
\hline
		&	& \\[-1ex]
{\sf NP} 	& 26\%	& 96.4\% \\
\hline
		&	& \\[-1ex]
decision	& 86\%	& 98.6\% \\
marked		& 10\%	& 86.8\% \\
no decision	&  4\%	& 73.0\% \\
\hline
\hline
		&	& \\[-1ex]
{\sf PP} 	& 24\%	& 97.9\% \\
\hline
		&	& \\[-1ex]
decision	& 92\%	& 99.2\% \\
marked		&  6\%	& 85.8\% \\
no decision	&  2\%	& 75.5\% \\
\hline
\hline
		&	& \\[-1ex]
{\em others}	& 18\%	& 94.7\% \\
\hline
		&	& \\[-1ex]
decision	& 91\%	& 98.0\% \\
marked		&  6\%	& 82.8\% \\
no decision	&  3\%	& 22.1\% \\
\hline
\hline
\end{tabular}
\end{center}
\hrule
\end{table}

\subsection{Error Analysis for Function Assignment}

The inspection of tagging errors reveals several sources of wrong
assignments. Table \ref{tab:errors} shows the 10 most frequent
errors\footnote{See appendix \ref{sec:tagsets} for a description of tags
used in the table.} which constitute 25\% of all errors (1509
errors occurred during 10 test runs).

Read the table in the following way: line 2 shows the second-most
frequent error. It concerns {\sf NP}s occurring in a sentence ({\sf S});
this combination occurred 1477 times during testing. In 286 of
these occurrences the {\sf NP} is assigned the grammatical function {\sf
OA} (accusative object) manually, but of these 286 cases the tagger
assigned the function {\sf SB} (subject) 56 times.

\begin{table}
\caption{The 10 most frequent errors in assigning grammatical functions.
The table shows a mother and a daughter node category, the frequency of
this particular combination (sum over 10 test runs), the grammatical
function assigned manually (and its frequency) and the grammatical
function assigned by the tagger (and its frequency).}
\label{tab:errors}
\hrule
\begin{center}\small
\begin{tabular}{@{\hspace{0pt}}r|l@{\hspace*{1pt}}l@{\hspace{0pt}}r|lr|lr}
&{\footnotesize phrase}	&\multicolumn{1}{c}{elem}&\multicolumn{1}{c|}{f}& \multicolumn{2}{c|}{original}&\multicolumn{2}{c}{assigned} \\
\hline
   &&&&&&&\\[-1ex]
1. & {\sf S}	& {\sf NP}	& 1477 	& {\sf SB}	&   894	& {\sf OA}	&  65 \\
2. & {\sf S}	& {\sf NP}	& 1477	& {\sf OA}	&   286	& {\sf SB}	&  56 \\
3. & {\sf NP}	& {\sf PP}	&  470	& {\sf PG}	&    52	& {\sf MNR}	&  50 \\
4. & {\sf S}	& {\sf VP}	&  613	& {\sf PD}	&    47 & {\sf OC}	&  42 \\
5. & {\sf PP}	& {\sf PP}	&  252	& {\sf PG}	&    30 & {\sf MNR}	&  30 \\
6. & {\sf VP}	& {\sf NP}	&  286	& {\sf DA}	&    32	& {\sf OA}	&  26 \\
7. & {\sf S}	& {\sf NP}	& 1477	& {\sf PD}	&    72 & {\sf SB}	&  25 \\
8. & {\sf S}	& {\sf NP}	& 1477	& {\sf MO}	&    33 & {\sf SB}	&  21 \\
9. & {\sf S}	& {\sf S}	&  186	& {\sf MO}	&    78 & {\sf PD}	&  21 \\
10.& {\sf VP}	& {\sf PP}	&  453	& {\sf SBP}	&    21 & {\sf MO}	&  21 \\
\end{tabular}
\end{center}
\hrule
\end{table}

The errors fall into the following classes:

\begin{enumerate}
\item[1.] There is insufficient information in the node labels to
	disambiguate the grammatical function.
\end{enumerate}
Line 1 is an example for insufficient information.
The tag {\sf NP} is uninformative about its case and
therefore the tagger has to distinguish {\sf SB}
(subject) and {\sf OA} (accusative object) on the basis of its
position, which is not very reliable in German. 
Missing information in the labels is the main source
of errors. Therefore, we currently investigate the benefits of a
morphological component and percolation of selected information
to parent nodes.

\begin{enumerate}
\item[2.] Due to the n-gram approach, the tagger only sees a
	local window of the sentences.
\end{enumerate}
Some linguistic knowledge is inherently global, e.g., there is at most
one subject in a sentence and one head in a {\sf VP}. Errors of this
type may be reduced by introducing finite state constraints that
restrict the possible sequences of functions within each phrase.

\begin{enumerate}
\item[3.] The manual annotation is wrong, and a correct
	tagger prediction is counted as an error.
\end{enumerate}
At earlier stages of annotation, the main source of errors was wrong
or missing manual annotation. In some cases, the tagger was able to
abstract from these errors during the training phase and subsequently
assigned the correct tag for the test data. However, when performing a
comparison against the corpus, these differences are marked as
errors. Most of these errors were eliminated by comparing two
independent annotations and cleaning up the data.

\subsection{Phrase Categories}

In this experiment, the reliability of assigning phrase categories given
the categories of the daughter nodes (they are supplied by the
annotator) was tested.

Consider the sentence in figure \ref{fig:ex}: two phrase categories
are to be determined ({\sf VP} and {\sf S}). The information given for
{\em selbst besucht Sabine} is the sequence of categories: adverb
({\sf ADV}), past participle ({\sf VVPP}), and proper noun ({\sf
NE}). The task is to assign category {\sf VP}. Subsequently, {\sf S}
is to be assigned based on the categories of the daughters {\sf VP},
{\sf VAFIN}, {\sf NE}, and {\sf ADV}.

\begin{table}
\caption{Levels of reliability and the percentage of cases in which the
tagger assigned a correct phrase category (or would have assigned if a
decision is forced).}
\label{tab:resphrase}
\hrule
\begin{center}
\begin{tabular}{l|rr}
	& cases		& correct \\
\hline
	& 		& 	\\[-1ex]
reliable	& 79\%	& 98.5\% \\
marked		& 16\%	& 90.4\% \\
unreliable	& 5\%	& 65.9\% \\
\hline
overall 	& 100\% & 95.4\% \\
\end{tabular}
\end{center}
\hrule
\end{table}

\begin{table}
\caption{Tagging accuracy for assigning phrase categories, depending on
the manually assigned category. For each category, the first row shows
the percentage of phrases belonging to a specific category (according to
manual assignment) and the percentage of correct assignments. The
following rows show the relative percentage and accuracy for different
levels of reliability.}
\label{tab:phrasecat}
\hrule
\begin{center}
\begin{tabular}{l|rr}
	& cases		& correct \\
\hline
\hline
		&	& \\[-1ex]
{\sf S} 	& 20\%	& 97.5\% \\
\hline
		&	& \\[-1ex]
decision	& 96\%	& 99.7\% \\
marked		&  2\%	& 63.2\% \\
no decision	&  2\%	& 29.0\% \\
\hline
\hline
		&	& \\[-1ex]
{\sf VP} 	& 9\%	& 93.2\% \\
\hline
		&	& \\[-1ex]
decision	& 71\%	& 96.4\%  \\
marked		& 24\%	& 91.3\% \\
no decision	& 5\%	& 60.9\% \\
\hline
\hline
		&	& \\[-1ex]
{\sf NP} 	& 29\%	& 96.1\% \\
\hline
		&	& \\[-1ex]
decision	& 81\%	& 99.3\% \\
marked		& 13\%	& 91.8\% \\
no decision	& 6\%	& 64.9\% \\
\hline
\hline
		&	& \\[-1ex]
{\sf PP} 	& 24\%	& 98.7\% \\
\hline
		&	& \\[-1ex]
decision	& 94\%	& 99.6\% \\
marked		& 4\%	& 92.5\% \\
no decision	& 2\%	& 70.8\% \\
\hline
\hline
		&	& \\[-1ex]
{\em others}	& 18\%	& 89.0\% \\
\hline
		&	& \\[-1ex]
decision	& 42\%	& 91.7\% \\
marked		& 45\%	& 90.6\% \\
no decision	& 12\%	& 73.2\% \\
\hline
\hline
\end{tabular}
\end{center}
\hrule
\end{table}

The extended tagger using a combined model as described in section
\ref{sec:tagphrase} was applied. 

Again, the corpus is divided into two disjoint parts, one for training
(90\% of the corpus), and one for testing (10\%). The procedure is
repeated 10 times with different partitions. Then the average accuracy
was calculated.

The same thresholds for search beams as for the first set of
experiments were used. 

Table \ref{tab:resphrase} shows tagging accuracy depending on the three
different levels of reliability.

Table \ref{tab:phrasecat} shows tagging accuracy depending on the
category of the phrase and the level of reliability. The table
contains the following information: the percentage of occurrences of
the particular phrase, the overall accuracy for that phrasal category
and the accuracy for each of the three reliability intervals.

\subsection{Error Analysis for Category Assignment}

When forced to make a decision (even in unreliable cases) 435 errors
occured during the 10 test runs (4.5\% error rate). Table
\ref{tab:phraseerr} shows the 10 most-frequent errors which constitute
50\% of all errors.

The most frequent error was the confusion of {\sf S} and {\sf VP}. They
differ in that sentences {\sf S} contain finite verbs and verb phrases {\sf VP}
contain non-finite verbs. But the tagger is trained on data that
contain incomplete sentences and therefore sometimes erroneously
assumes an incomplete {\sf S} instead of a {\sf VP}. To avoid this type
of error, the tagger should be able to take the neighborhood of phrases
into account. Then, it could detect the finite verb that completes the
sentence.

Adjective phrases {\sf AP} and noun phrases {\sf NP} are confused by the
tagger (line 5 in table \ref{tab:phraseerr}), since almost all {\sf
AP}'s can be {\sf NP}'s. This error could also be fixed by inspecting
the context and detecting the associated {\sf NP}.

As for assigning grammatical functions, insufficient information in
the labels is a significant source of errors, cf. the second-most
frequent error. A large number of cardinal-noun pairs forms a
numerical component ({\sf NM}), like {\em 7 Millionen}, {\em 50
Prozent}, etc ({\em 7 million, 50 percent}). But this combination also
occurs in {\sf NP}s like {\em 20 Leute}, {\em 3 Monate}, \dots ({\em
20 people}, {\em 3 months}), which are mis-tagged since they are less
frequent. This can be fixed by introducing an extra tag for nouns
denoting numericals.

\begin{table}
\bigskip
\caption{The 10 most frequent errors in assigning phrase categories
(summed over reliability levels). The table shows the phrase category 
assigned manually (and its frequency) and the category erroneously
assigned by the tagger (and its frequency).}
\label{tab:phraseerr}
\hrule
\begin{center}\small
\begin{tabular}{@{\hspace{0pt}}r|l@{\hspace{0pt}}r|l@{\hspace{0pt}}r}
&{\footnotesize phrase}	&\multicolumn{1}{c|}{f}&
\multicolumn{1}{c}{assigned} & \multicolumn{1}{c}{f} \\
\hline
   &		&	&		&	\\[-1ex]
1. & {\sf VP}	&  828 	& {\sf S}	&    46	\\
2. & {\sf NP}	& 2812	& {\sf NM}	&    32	\\
3. & {\sf NP}	& 2812	& {\sf PP}	&    31	\\
4. & {\sf NP}	& 2812	& {\sf S}	&    25 \\
5. & {\sf AP}	&  419	& {\sf NP}	&    15 \\
6. & {\sf DL}	&   20	& {\sf CS}	&    15	\\
7. & {\sf PP}	& 2298	& {\sf NP}	&    15 \\
8. & {\sf S}	& 1910	& {\sf NP}	&    15 \\
9. & {\sf AP}	&  419	& {\sf PP}	&    11 \\
10.& {\sf MPN}	&  293	& {\sf NP}	&    11 \\
\end{tabular}
\end{center}
\hrule
\end{table}

\section{Conclusion}

A German newspaper corpus is currently being annotated with a new
annotation scheme especially designed for free word order languages.

Two levels of automatic annotation (level 1: assigning grammatical
functions and level 2: assigning phrase categories) have been
presented and evaluated in this paper.

The overall accuracy for assigning grammatical functions is 94.2\%,
ranging from 89\% to 98\%, depending on the type of phrase. The least
accuracy is achieved for sentences, the best for prepositional
phrases.  By suppressing unreliable decisions, precision can be
increased to range from 92\% to 99\%.

The overall accuracy for assigning phrase categories is 95.4\%,
ranging from 89\% to 99\%, depending the category. By suppressing
unreliable decisions, precision can also be increased to range from
92\% to over 99\%.

In the error analysis, the following sources of misinterpretation
could be identified: insufficient linguistic information in the nodes
(e.g., missing case information), and insufficient information about
the global structure of phrases (e.g., missing valency
information). Morphological information in the tagset, for example,
helps to identify the objects and the subject of a sentence.  Using a
more fine-grained tagset, however, requires methods for adjusting the
granularity of the tagset to the size (and coverage) of the corpus, in
order to cope with the sparse data problem.

\section{Acknowledgements}

This work is part of the DFG Sonderforschungsbereich 378 {\em
Resource-Adaptive Cognitive Processes}, Project C3 {\em Concurrent
Grammar Processing}.

We wish to thank the universities of Stuttgart and  T\"ubingen for kindly
providing us with a hand-corrected part-of-speech tagged  corpus. We
also wish to thank Jason Eisner, Robert MacIntyre and Ann Taylor for
valuable discussions on dependency parsing and the Penn Treebank
annotation. Special thanks go to Oliver Plaehn, who implemented the
annotation tool, and to our six fearless annotators.

\section*{Appendix A: Tagsets}

\makeatletter
\write\@auxout{\string
  \newlabel{sec:tagsets}{{A}{\thepage}}}
\makeatother

This section contains descriptions of tags used in this
paper. These are {\em not} complete lists.

\subsection*{A.1 Part-of-Speech Tags}

We use the Stuttgart-T{\"u}bingen-Tagset. The complete set is described
in \cite{Schiller;Thielen:94}.

\noindent
\begin{tabular}{ll}
{\sf ADJA}      & attributive adjective\\
{\sf ADJD}      & adverbial adjective\\
{\sf ADV}       & adverb \\
{\sf APPR}      & preposition\\
{\sf ART}       & article\\
{\sf CARD}	& cardinal number\\
{\sf FM}	& foreign material \\
{\sf KOKOM}	& comparing conjunction\\
{\sf KOUS}	& sub-ordinating conjunction \\
{\sf NE}	& proper noun \\
{\sf NN}        & common noun \\
{\sf PIAT}	& indefinite pronoun\\
{\sf PPER}      & personal pronoun\\
{\sf PTKNEG}    & negation\\
{\sf VAFIN}     & finite auxiliary \\
{\sf VMFIN}	& finite modal verb\\
{\sf VVPP}      & past participle of main verb\\
\end{tabular}

\subsection*{A.2 Phrasal Categories}

\noindent
\begin{tabular}{ll}
{\sf AP}	& adjective phrase\\
{\sf CS}	& coordination of sentences\\
{\sf DL}	& discurse level\\
{\sf MPN}	& multi-word proper noun \\
{\sf NM}	& multi-token numerical\\
{\sf NP}	& noun phrase\\
{\sf PP}        & prepositional phrase \\
{\sf S}         & sentence \\
{\sf VP}        & verb phrase \\

\end{tabular}

\subsection*{A.3 Grammatical Functions}

\noindent
\begin{tabular}{ll}
{\sf AC}	& adpositional case marker\\
{\sf CJ}	& conjunct\\
{\sf DA}	& dative\\
{\sf HD}	& head\\
{\sf JU}        & junctor \\
{\sf MNR}       & post-nominal modifier\\
{\sf MO}        & modifier \\
{\sf NG}	& negation\\
{\sf NK}        & noun kernel\\
{\sf OA}        & accusative object \\
{\sf OC}        & clausal object \\
{\sf PD}        & predicative \\
{\sf PG}        & pseudo genitive \\
{\sf PNC}       & proper noun component \\
{\sf SB}        & subject \\
{\sf SBP}	& passivized subject \\
{\sf SP}        & subject or predicative \\

\end{tabular}


\begin{thebibliography}{}

\bibitem[\protect\citename{Abney}1996]{Abney96}
Abney, Steven.
\newblock 1996.
\newblock Partial parsing via finite-state cascades.
\newblock In {\em Proceedings of the ESSLLI'96 Robust Parsing Workshop},
  Prague, Czech Republic.

\bibitem[\protect\citename{Black \bgroup et al.\egroup }1996]{Black96}
Black, Ezra, Stephen Eubank, Hideki Kashioka, David Magerman, Roger Garside,
  and Geoffrey Leech.
\newblock 1996.
\newblock Beyond skeleton parsing: Producing a comprehensive large-scale
  general-english treebank with full grammaticall analysis.
\newblock In {\em Proc.\ of COLING-96}, pages 107--113, Kopenhagen, Denmark.

\bibitem[\protect\citename{Brown \bgroup et al.\egroup }1992]{Brown92}
Brown, P.~F., V.~J.~Della Pietra, Peter~V. deSouza, Jenifer~C. Lai, and
  Robert~L. Mercer.
\newblock 1992.
\newblock Class-based $n$-gram models of natural language.
\newblock {\em Computational Linguistics}, 18(4):467--479.

\bibitem[\protect\citename{Collins}1996]{Collins96}
Collins, Michael.
\newblock 1996.
\newblock A new statistical parser based on bigram lexical dependencies.
\newblock In {\em Proceedings of ACL-96}, Sant Cruz, CA, USA.

\bibitem[\protect\citename{Cutting \bgroup et al.\egroup }1992]{Cutting92}
Cutting, Doug, Julian Kupiec, Jan Pedersen, and Penelope Sibun.
\newblock 1992.
\newblock A practical part-of-speech tagger.
\newblock In {\em Proceedings of the 3rd Conference on Applied Natural Language
  Processing (ACL)}, pages 133--140.

\bibitem[\protect\citename{Eisner}1996]{Eisner96}
Eisner, Jason~M.
\newblock 1996.
\newblock Three new probabilistic models for dependency parsing: An
  exploration.
\newblock In {\em Proceedings of COLING-96}, Kopenhagen, Denmark.

\bibitem[\protect\citename{Feldweg}1995]{Feldweg95}
Feldweg, Helmut.
\newblock 1995.
\newblock Implementation and evaluation of a german hmm for pos disambiguation.
\newblock In {\em Proceedings of EACL-SIGDAT-95 Workshop}, Dublin, Ireland.

\bibitem[\protect\citename{Lehmann \bgroup et al.\egroup }1996]{LehmannEA:96}
Lehmann, Sabine, Stephan Oepen, Sylvie Regnier-Prost, Klaus Netter, Veronika
  Lux, Judith Klein, Kirsten Falkedal, Frederik Fouvry, Dominique Estival, Eva
  Dauphin, Herv{\'e} Compagnion, Judith Baur, Lorna Balkan, and Doug Arnold.
\newblock 1996.
\newblock {{\sc tsnlp} --- Test Suites for Natural Language Processing}.
\newblock In {\em Proceedings of COLING 1996}, Kopenhagen.

\bibitem[\protect\citename{Marcus \bgroup et al.\egroup }1994]{Marcusea:94}
Marcus, Mitchell, Grace Kim, Mary~Ann Marcinkiewicz, Robert MacIntyre, Ann
  Bies, Mark Ferguson, Karen Katz, and Britta Schasberger.
\newblock 1994.
\newblock The penn treebank: Annotating predicate argument structure.
\newblock In {\em Proceedings of the Human Language Technology Workshop}, San
  Francisco, Morgan Kaufmann.

\bibitem[\protect\citename{Rabiner}1989]{Rabiner89}
Rabiner, L.~R.
\newblock 1989.
\newblock A tutorial on hidden markov models and selected applications in
  speech recognition.
\newblock In {\em Proceedings of the IEEE}, volume 77(2), pages 257--285.

\bibitem[\protect\citename{Sampson}1995]{Sampson95}
Sampson, Geoffrey.
\newblock 1995.
\newblock {\em English for the Computer}.
\newblock Oxford University Press, Oxford.

\bibitem[\protect\citename{Skut \bgroup et al.\egroup }1997]{Skutea97}
Skut, Wojciech, Brigitte Krenn, Thorsten Brants, and Hans Uszkoreit.
\newblock 1997.
\newblock An annotation scheme for free word order languages.
\newblock In {\em Proceedings of ANLP-97}, Washington, DC.

\bibitem[\protect\citename{Tapanainen and
  J{\"a}rvinen}1997]{TapanainenJaervinen:97}
Tapanainen, Pasi and Timo J{\"a}rvinen.
\newblock 1997.
\newblock A non-projective dependency parser.
\newblock In {\em Proceedings of ANLP-97}, Washington, DC.

\bibitem[\protect\citename{Thielen and Schiller}1995]{Schiller;Thielen:94}
Thielen, Christine and Anne Schiller.
\newblock 1995.
\newblock {Ein kleines und erweitertes Tagset f{\"u}rs Deutsche}.
\newblock In {\em Tagungsberichte des Arbeitstreffens Lexikon + Text 17./18.
  Februar 1994, Schlo{\ss} Hohent{\"u}bingen. Lexicographica Series Maior},
  T\"ubingen. Niemeyer.

\end{thebibliography}
\end{document}